\documentclass{article} % For LaTeX2e
\usepackage{iclr2017_conference,times}
\usepackage{hyperref}
\usepackage{url}
\usepackage{graphicx}
\usepackage{caption}
\usepackage{subcaption}
\usepackage{amsmath}

\newcommand{\stitle}[1]{\vspace{5pt} \noindent\textbf{#1.}\ }

\newtheorem{proposition}{Proposition}

\newcommand{\esm}[1]{\ensuremath{#1}}

\newcommand{\ms}[1]{\esm{\mathsf{#1}}}

\newcommand\reals{\ms{R}} \newcommand\incp{\ms{Incp}}
 \newcommand\img{\ms{img}}
\newcommand\pimportance{\mathcal{P}}
\newcommand\molfeatures{\ms{Features}}

\newcommand\grad{\bigtriangledown}
\newcommand\sparam{\alpha}
\newcommand\im{\ms{img}}
\newcommand\mol{\ms{mol}}
\newcommand\pathfn{\gamma}

\newcommand\synteq{::=}
\newcommand\intgrads{\ms{InteriorGrads}}
\newcommand\integratedgrads{\ms{IntegratedGrads}}

\newcommand\intpimportance{\ms{InteriorPixelImportance}}
\title{Gradients of Counterfactuals}

\author{Mukund Sundararajan, Ankur Taly \& Qiqi Yan \\
Google Inc.\\
Mountain View, CA 94043, USA
\\ \texttt{\{mukunds,ataly,qiqiyan\}@google.com} \\ }

\begin{document}

\maketitle

\begin{abstract}
  Gradients have been used to quantify feature importance in machine
  learning models.  Unfortunately, in nonlinear deep networks, not only
  individual neurons but also the whole network can saturate, and as a
  result an
  important input feature can have a tiny gradient. We study various
  networks, and observe that this phenomena is indeed widespread,
  across many inputs.

  We propose to examine \emph{interior gradients}, which are gradients
  of counterfactual inputs constructed by scaling down the
  original input. We apply our method to the GoogleNet architecture
  for object recognition in images, as well as a ligand-based virtual
  screening network with categorical features and an LSTM based language
  model for the Penn Treebank dataset. We visualize how
  interior gradients better capture feature importance.  Furthermore,
  interior gradients are applicable to a wide variety of
  deep networks, and have the \emph{attribution} property that
  the feature importance scores sum to the the prediction score.

  Best of
  all, interior gradients can be computed \emph{just as easily as}
  gradients. In contrast, previous methods are complex to implement,
  which hinders practical adoption.
\end{abstract}

\section{Introduction}

Practitioners of machine learning regularly inspect the coefficients
of linear models as a measure of feature importance. This process
allows them to understand and debug these models. The natural analog
of these coefficients for deep models are the gradients of the
prediction score with respect to the input. For linear models, the
gradient of an input feature is equal to its coefficient. For deep
nonlinear models, the gradient can be thought of as a local linear
approximation (\cite{SVZ13}). Unfortunately, (see the next section),
the network can saturate and as a result an important input
feature can have a tiny gradient.

While there has been other work (see Section~\ref{sec:related}) to
address this problem, these techniques involve instrumenting the
network. This instrumentation currently involves significant developer
effort because they are not primitive operations in standard machine
learning libraries. Besides, these techniques are not simple to
understand---they invert the operation of the network in different
ways, and have their own peculiarities---for instance, the feature
importances are not invariant over networks that compute the exact same
function (see Figure~\ref{fig:deeplift}).

In contrast, the method we propose builds on the very familiar,
primitive concept of the gradient---all it involves is inspecting the
gradients of a few carefully chosen counterfactual inputs that are
scaled versions of the initial input.  This allows anyone who knows
how to extract gradients---presumably even novice practitioners that
are not very familiar with the network's implementation---to
\emph{debug} the network. Ultimately, this seems essential to ensuring
that deep networks perform predictably when deployed.

\section{Our Technique}
\subsection{Gradients Do Not Reflect Feature Importance}
Let us start by investigating the performance of gradients as a measure of feature importance.
We use an object recognition network built using the GoogleNet
architecture (\cite{SLJSRAEVR14}) as a running example; we refer to this
network by its codename Inception. (We present applications of our
techniques to other networks in Section~\ref{sec:beyond-inception}.)
The network has been trained on the ImageNet object recognition
dataset (\cite{ILSVRC15}). It is is $22$ layers deep with a softmax layer on top for
classifying images into one of the $1000$ ImageNet object classes.
The input to the
network is a $224 \times 224$ sized RGB image.

Before evaluating the use of gradients for feature importance, we introduce some basic notation that is used throughout
the paper.

We represent a $224\times 224$ sized RGB image as a vector in
$\reals^{224\times 224\times 3}$. Let $\incp^L: \reals^{224\times 224\times 3} \rightarrow [0,1]$ be
the function represented by the Inception network that computes the
softmax score for the object class labeled $L$.
%\footnote{While we explain the technique with respect to the
%  highest scoring object class, it can be applied to any specific
%  object class}
Let $\grad\incp^L(\img)$ be the gradients of $\incp^L$
 at the input
image $\img$. Thus, the vector $\grad\incp^L(\img)$ is the same size as
the image and lies in $\reals^{224\times 224\times 3}$.  As a shorthand, we write
$\grad\incp^L_{i,j,c}(\img)$ for the gradient of a specific pixel
$(i,j)$ and color channel $c \in \{R,G,B\}$.

\begin{figure}
  \begin{center}
    \begin{subfigure}{.9\textwidth}
      \centering
      \includegraphics[width=0.9\textwidth]{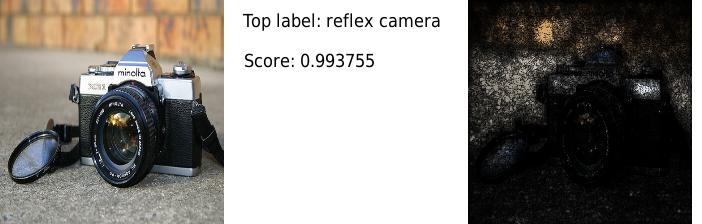}
      \caption{Original image.}\label{fig:camera-finalgrad}
    \end{subfigure}
    \begin{subfigure}{.9\textwidth}
      \centering
      \includegraphics[width=0.9\textwidth]{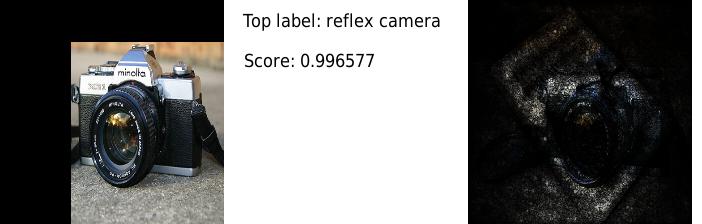}
      \caption{Ablated image.}\label{fig:camera-abl-finalgrad}
    \end{subfigure}
    %\fbox{\rule[-.5cm]{0cm}{4cm} \rule[-.5cm]{4cm}{0cm}}
  \end{center}
  \caption{Pixel importance using gradients at the image.}
\end{figure}

We compute the gradients of $\incp^L$ (with respect to the image) for the
highest-scoring object class, and then aggregate the gradients $\grad\incp^L(\img)$
along the color dimension to obtain pixel importance scores.\footnote{
  These pixel importance scores are similar to the gradient-based
  saliency map defined by~\cite{SVZ13} with the
  difference being in how the gradients are aggregated along the color
  channel.}
\begin{equation}\label{eq:pimportance}
\forall i, j: \pimportance^L_{i,j}(\img)  ~\synteq~  \Sigma_{c \in \{R,G,B\}} \vert\grad\incp^L_{i,j,c}(\img)\vert
\end{equation}
Next, we visualize pixel importance scores by scaling the intensities of the
pixels in the original image in proportion to their respective scores;
thus, higher the score brighter would be the pixel. Figure
\ref{fig:camera-finalgrad} shows a visualization for an
image for which the highest scoring object class is ``reflex camera''
with a softmax score of $0.9938$.

Intuitively, one would expect the the high gradient pixels for this
classification to be ones falling on the camera or those providing
useful context for the classification (e.g., the lens cap). However,
most of the highlighted pixels seem to be on the left or above the
camera, which to a human seem not essential to the prediction. This could
either mean that (1) the highlighted pixels are somehow important for
the internal computation performed by the Inception network, or (2)
gradients of the image fail to appropriately quantify pixel
importance.

Let us consider hypothesis (1). In order to test it we ablate parts of
the image on the left and above the camera (by zeroing out the pixel
intensities) and run the ablated image through the Inception
network. See Figure~\ref{fig:camera-abl-finalgrad}. The top
predicted category still remains ``reflex camera'' with a softmax
score of $0.9966$ --- slightly higher than before. This indicates that
the ablated portions are indeed irrelevant to the classification. On
computing gradients of the ablated image, we still find that most of
the high gradient pixels lie outside of the camera.  This suggests
that for this image,  it is in fact hypothesis (2) that holds true.
Upon studying more images (see Figure~\ref{fig:intgrad-finalgrad}), we find that
the gradients often fail to highlight the relevant pixels for the predicted
object label.

\subsection{Saturation}
In theory, it is easy to see that the gradients may not reflect
feature importance if the prediction function flattens in the vicinity
of the input, or equivalently, the gradient of the prediction function
with respect to the input is tiny in the vicinity of the input vector. This
is what we call \emph{saturation}, which has also been reported in previous
work (\cite{SGSK16}, \cite{GB10}).

We analyze how widespread saturation is in the Inception network by
inspecting the behavior of the network on \textbf{counterfactual images}
obtained by uniformly scaling pixel intensities from zero to their
values in an actual image. Formally, given an input image $\im \in \reals^{224\times 224\times 3}$,
the set of counterfactual images is
\begin{equation}
\{\sparam~\im~\vert~ 0 \leq \sparam \leq 1\}
\end{equation}
Figure~\ref{fig:sat-incp-smax} shows the trend in the softmax output of the
highest scoring class, for thirty randomly chosen images form the ImageNet
dataset. More specifically, for each image $\im$, it shows the trend in
$\incp^L(\sparam~\im)$ as $\sparam$ varies from zero to one with $L$ being
the label of highest scoring object class for $\im$. It is easy to
see that the trend flattens (saturates) for all images
$\sparam$ increases. Notice that saturation is present even for images whose
final score is significantly below $1.0$. Moreover, for a majority of images,
saturation happens quite soon when $\sparam=0.2$.

One may argue that since the output of the Inception network is the result of
applying the softmax function to a vector of activation values,
the saturation is expected due to the squashing property of the softmax function.
However, as shown in Figure~\ref{fig:sat-incp-presmax}, we find that even the
pre-softmax activation scores for the highest scoring class saturate.

In fact, to our surprise, we found that the saturation is inherently present
in the Inception network and the outputs of the intermediate layers also
saturate. We plot the distance between the intermediate layer neuron activations
for a scaled down input image and the actual input image with respect to the
scaling parameter, and find that the trend flattens. Due to lack of space, we
provide these plots in Figure~\ref{fig:sat-incp-intermediate} in the appendix.

It is quite clear from these plots that saturation is widespread
across images in the Inception network, and there is a lot more
activity in the network for counterfactual images at relatively low
values of the scaling parameter $\sparam$. This observation forms the basis of our
technique for quantifying feature importance.

Note that it is well known that the saturation of gradients prevent
the model from converging to a good quality minima (\cite{GB10}). So one
may expect good quality models to not have saturation and hence for
the (final) gradients to convey feature importance. Clearly, our observations on the
Inception model show that this is not the case. It has good prediction
accuracy, but also exhibits saturation (see Figure~\ref{fig:sat-incp}).
Our hypothesis is that the gradients of important features are
\emph{not} saturated early in the training process. The gradients only
saturate \emph{after} the features have been learned adequately, i.e.,
the input is far away from the decision boundary.

\subsection{Interior Gradients}
We study the importance of input features in a prediction made for an
input by examining the gradients of the counterfactuals obtained by
scaling the input; we call this set of gradients \textbf{interior gradients}.

While the method of examining gradients of counterfactual inputs is
broadly applicable to a wide range of networks, we first explain it in
the context of Inception. Here, the counterfactual image inputs we consider
are obtained by uniformly scaling pixel intensities from zero to their
values in the actual image (this is the same set of counterfactuals
that was used to study saturation). The interior gradients are the
gradients of these images.
\begin{equation}
  \intgrads(\img) ~\synteq~ \{\grad \incp(\sparam~\img)~\vert~ 0 \leq \sparam \leq 1\}
\end{equation}

\begin{figure}[!htb]
  \begin{center}
    \begin{subfigure}{.5\textwidth}
      \centering
      \includegraphics[width=0.8\textwidth]{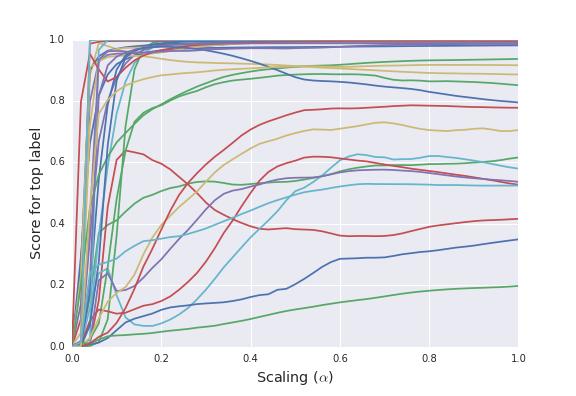}
      \caption{Softmax score for top label}\label{fig:sat-incp-smax}
    \end{subfigure}%
    \begin{subfigure}{.5\textwidth}
      \centering
      \includegraphics[width=0.8\textwidth]{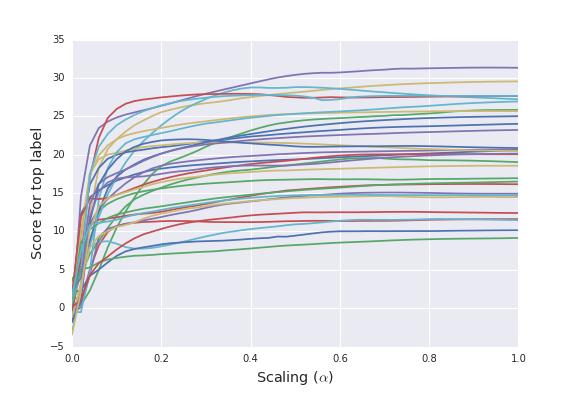}
      \caption{Pre-softmax score for top label}\label{fig:sat-incp-presmax}
    \end{subfigure}
  %\fbox{\rule[-.5cm]{0cm}{4cm} \rule[-.5cm]{4cm}{0cm}}
  \end{center}
  \caption{Saturation in Inception}\label{fig:sat-incp}
\end{figure}

These interior gradients explore the behavior of the network along the
entire scaling curve depicted in Figure~\ref{fig:sat-incp-smax}, rather
than at a specific point. We can aggregate the interior gradients along
the color dimension to obtain interior pixel importance scores using
equation~\ref{eq:pimportance}. 
\begin{equation}        
  \intpimportance(\img) ~\synteq~ \{\pimportance(\sparam~\img)~\vert~0 \leq \sparam \leq 1\}
\end{equation}

We individually visualize the pixel importance scores for each scaling
parameter $\sparam$ by scaling the intensities of the pixels in
the actual image in proportion to their scores. The visualizations show
how the importance of each pixel evolves as we scale the image,
with the last visualization being identical to one generated by gradients
at the actual image. In this regard, the interior gradients offer strictly
more insight into pixel importance than just the gradients at the actual image.

\begin{figure}
  \centering
  \begin{subfigure}{0.8\textwidth}
    \centering
    \includegraphics[width=0.9\textwidth]{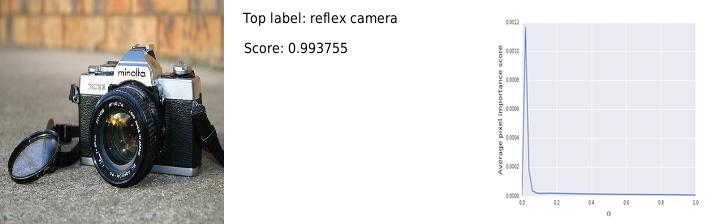}
    \caption*{Input image and trend of the pixel importance scores
      obtained from interior gradients.}
    \vspace{0.5cm}
  \end{subfigure}
  \begin{subfigure}{.2\textwidth}
    \centering
    \includegraphics[width=0.9\textwidth]{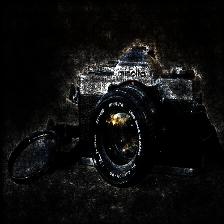}
    \caption*{$\alpha= 0.02$}
  \end{subfigure}%
  \begin{subfigure}{.2\textwidth}
    \centering
    \includegraphics[width=0.9\textwidth]{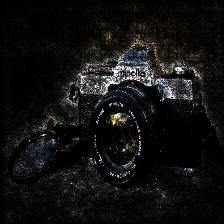}
    \caption*{$\alpha= 0.04$}
  \end{subfigure}%
  \begin{subfigure}{.2\textwidth}
    \centering
    \includegraphics[width=0.9\textwidth]{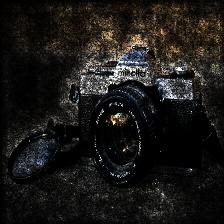}
    \caption*{$\alpha= 0.06$}
  \end{subfigure}%
  \begin{subfigure}{.2\textwidth}
    \centering
    \includegraphics[width=0.9\textwidth]{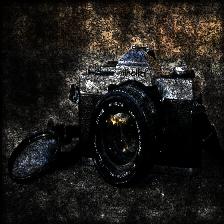}
    \caption*{$\alpha= 0.08$}
  \end{subfigure}%
  \begin{subfigure}{.2\textwidth}
    \centering
    \includegraphics[width=0.9\textwidth]{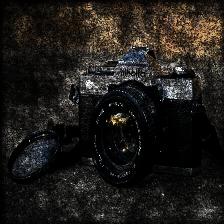}
    \caption*{$\alpha= 0.1$}
  \end{subfigure}
  \begin{subfigure}{.2\textwidth}
    \centering
    \includegraphics[width=0.9\textwidth]{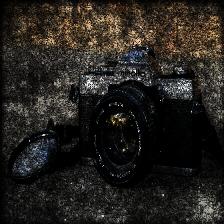}
    \caption*{$\alpha= 0.2$}
  \end{subfigure}%
  \begin{subfigure}{.2\textwidth}
    \centering
    \includegraphics[width=0.9\textwidth]{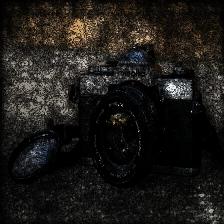}
    \caption*{$\alpha= 0.4$}
  \end{subfigure}%
  \begin{subfigure}{.2\textwidth}
    \centering
    \includegraphics[width=0.9\textwidth]{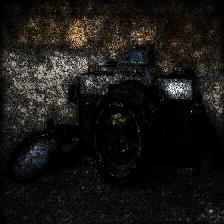}
    \caption*{$\alpha= 0.6$}
  \end{subfigure}%
  \begin{subfigure}{.2\textwidth}
    \centering
    \includegraphics[width=0.9\textwidth]{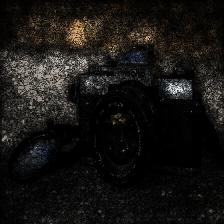}
    \caption*{$\alpha= 0.8$}
  \end{subfigure}%
  \begin{subfigure}{.2\textwidth}
    \centering
    \includegraphics[width=0.9\textwidth]{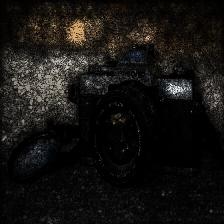}
    \caption*{$\alpha= 1.0$}
  \end{subfigure}
  \caption{\textbf{Visualization of interior gradients.} Notice that the visualizations at
    lower values of the scaling parameter ($\alpha$) are sharper and much better at surfacing
    important features of the input image.}\label{fig:camera-intgrads}
\end{figure}

Figure~\ref{fig:camera-intgrads} shows the visualizations for the ``reflex camera'' image
from Figure ~\ref{fig:camera-finalgrad} for various values of the scaling parameter
$\sparam$. The plot in the top right corner shows the trend in the absolute magnitude of the
average pixel importance score. The magnitude is significantly larger
at lower values of $\sparam$ and nearly zero at higher values --- the latter
is a consequence of saturation.
Note that each visualization is only indicative of the relative distribution of
the importance scores across pixels and not the absolute magnitude of the scores, i.e.,
the later snapshots are responsible for tiny increases in the scores as the chart in the top right depicts.

The visualizations show that at lower values of $\sparam$, the pixels that lie
on the camera are most important, and as $\sparam$ increases, the region above
the camera gains importance. Given the high magnitude of gradients at
lower values of $\sparam$, we consider those gradients to be the primary
drivers of the final prediction score. They are more indicative of
feature importance in the prediction compared to the gradients at the
actual image (i.e., when $\sparam = 1$).

The visualizations of the interior pixel gradients can also be viewed
together as a single animation that chains the visualizations in sequence
of the scaling parameter.  This animation offers a concise yet complete
summary of how pixel importance moves around the image as the scaling
parameter increase from zero to one.

\stitle{Rationale}
While measuring saturation via counterfactuals seems natural, using
them for quantifying feature importance deserves some discussion. The
first thing one may try to identify feature importance is to examine
the deep network like one would with human authored code. This seems
hard; just as deep networks employ distributed
representations (such as embeddings), they perform convoluted (pun
intended) distributed reasoning. So instead, we choose to probe the
network with several counterfactual inputs (related to the input at
hand), hoping to trigger all the internal workings of the network.
This process would help summarize the effect of the network on the
protagonist input; the assumption being that the input is human
understandable. Naturally, it helps to work with
gradients in this process as via back propagation, they induce an
aggregate view over the function computed by the neurons.

Interior gradients use counterfactual inputs to artifactually induce a
procedure on how the network’s attention moves across the image as it
compute the final prediction score. From the animation, we gather that the
network focuses on strong and distinctive patterns in the image at
lower values of the scaling parameter, and subtle and weak patterns in
the image at higher values.  Thus, we speculate that the network's
computation can be loosely abstracted by a procedure that first
recognize distinctive features of the image to make an initial
prediction, and then fine tunes (these are small score jumps as
the chart in Figure~\ref{fig:camera-intgrads} shows) the prediction
using weaker patterns in the image.
%%
%% Why this may be the case? When the scaling parameter is very small (or
%% equivalently the brightness is very low), only the distinctive
%% patterns in the image are visible. Consequently, the network only
%% focusses on them. As the image is scaled up, we speculate that two
%% things happen - (1) Subtle and less distinctive patterns gain
%% visibility, and (2) Saturation increases in the network.  We further
%% posit that neurons responsible for the distinctive features move into
%% saturation, thereby dropping the gradients for those features. This is
%% why the interior gradients at high values of the scaling parameter
%% depict seemingly insignificant patterns.
%

\subsection{Cumulating Interior Gradients}
A different summarization of the interior gradients can be obtained by
cumulating them. While there are a few ways of cumulating
counterfactual gradients, the approach we take has the nice
\emph{attribution} property (Proposition~\ref{prop:additivity}) that
the feature importance scores approximately add up to the prediction
score. The feature importance scores are thus also referred to as
\emph{attributions}.

Notice that the set of counterfactual images $\{\sparam~\im~\vert~ 0
\leq \sparam \leq 1\}$ fall on a straight line path in
$\reals^{224\times224\times3}$. Interior gradients --- which are the
gradients of these counterfactual images --- can be cumulated by
integrating them along this line. We call the resulting gradients as
\emph{\bf integrated gradients}. In what follows, we formalize
integrated gradients for an arbitrary function $F: \reals^n
\rightarrow [0,1]$ (representing a deep network), and an arbitrary set
of counterfactual inputs falling on a path in $\reals^n$.

Let $x \in \reals^n$ be the input at hand, and
$\pathfn = (\pathfn_1, \ldots, \pathfn_n): [0,1]
\rightarrow \reals^n$ be a smooth function specifying the set of
counterfactuals; here, $\pathfn(0)$ is the \emph{baseline} input (for
Inception, a black image), and $\pathfn(1)$ is the actual input (for
Inception, the image being studied). Specifically,
$\{\pathfn(\sparam)~\vert~0 \leq \sparam \leq 1\}$ is the set of
counterfactuals (for Inception, a series of images that interpolate
between the black image and the actual input).

The integrated gradient along the $i^{th}$ dimension for an input $x \in
\reals^n$ is defined as follows.
\begin{equation}
\integratedgrads_i(x) \synteq \int_{\sparam=0}^{1} \tfrac{\partial F(\pathfn(\sparam))}{\partial \pathfn_i(\sparam)  }~\tfrac{\partial \pathfn_i(\sparam)}{\partial \sparam}  ~d\sparam
\end{equation}
where $\tfrac{\partial F(x)}{\partial x_i}$ is the gradient of
$F$ along the $i^{th}$ dimension at $x$.

A nice technical property of the integrated gradients is that they add
up to the difference between the output of $F$ at the final
counterfactual $\pathfn(1)$ and the \emph{baseline} counterfactual
$\pathfn(0)$.  This is formalized by the proposition below, which is
an instantiation of the fundamental theorem of calculus for path
integrals.
\begin{proposition}\label{prop:additivity}
  If $F: \reals^n \rightarrow \reals$ is differentiable almost
  everywhere
  \footnote{Formally, this means that the partial derivative of $F$
    along each input dimension satisfies Lebesgue's integrability
    condition, i.e., the set of discontinuous points has measure
    zero. Deep networks built out of Sigmoids, ReLUs, and pooling
    operators should satisfy this condition.}, and $\pathfn: [0,1]
  \rightarrow \reals^n$ is smooth then
  $$\Sigma_{i=1}^{n} \integratedgrads_i(x) = F(\pathfn(1)) -
  F(\pathfn(0))$$
\end{proposition}

For most deep networks, it is possible to choose counterfactuals such
that the prediction at the baseline counterfactual is near zero
($F(\pathfn(0)) \approx 0$).
%% \footnote{We did have trouble finding a
%%   baseline couterfactual for an RNN model that simulated the workings of
%%   a traffic light intersection between a main road and a side street;
%%   the naive benchmark counterfactual was one of no traffic at either
%%   intersection. But this did not have the lack of semantics that a
%%   black image or pure noise has for the Inception network in the sense
%%   that \emph{no} interesting labels are activated.}
For instance, for the
Inception network, the counterfactual defined by the scaling path
satisfies this property as $\incp(0^{224\times 224\times 3}) \approx
0$. In such cases, it follows from the Proposition that the integrated
gradients form an attribution of the prediction output $F(x)$, i.e.,
they almost exactly distribute the output to the individual input features.

The additivity property provides a form of sanity checking for the integrated
gradients and ensures that we do not under or over attribute to
features. This is a common pitfall for attribution schemes based on
feature ablations, wherein, an ablation may lead to small or a large change in
the prediction score depending on whether the ablated feature
interacts disjunctively or conjunctively to the rest of the features.
This additivity is even more desirable when the network’s score is
numerically critical, i.e., the score is not used purely in an ordinal
sense. In this case, the attributions (together with additivity)
guarantee that the attributions are in the \emph{units} of the score,
and account for all of the score.

We note that these path integrals of gradients have been used to perform
attribution in the context of small non-linear polynomials
(\cite{SS11}), and also within the cost-sharing literature in economics
where function at hand is a cost function that models the cost of a
project as a function of the demands of various participants, and the
attributions correspond to cost-shares. The specific “path” we use
corresponds to a cost-sharing method called Aumann-Shapley (\cite{AS74}).

\stitle{Computing integrated gradients}
The integrated gradients can be efficiently approximated by Riemann
sum, wherein, we simply sum the gradients at points occurring at
sufficiently small intervals along the path of counterfactuals.
\begin{equation}
  \integratedgrads_i^{approx}(x) \synteq \Sigma_{k=1}^m  \tfrac{\partial F(\pathfn(k/m))}{\partial \pathfn_i(\sparam)  }~ (\pathfn(\tfrac{k}{m})-\pathfn(\tfrac{k-1}{m}))
\end{equation}
Here $m$ is the number of steps in the Riemman approximation of the integral.
Notice that the approximation simply involves computing the gradient in
a for loop; computing the gradient is central to deep learning and is a
pretty efficient operation. The implementation should therefore be
straightforward in most deep learning frameworks. For instance, in
TensorFlow (\cite{tensorflow}), it essentially amounts to calling
{\tt tf.gradients} in a loop over the set of counterfactual
inputs (i.e., $\pathfn(\tfrac{k}{m})~\mbox{for}~ k = 1, \ldots, m$), which
could also be batched. Going forward, we abuse the term ``integrated gradients''
to refer to the approximation described above.

\stitle{Integrated gradients for Inception}
We compute the integrated gradients for the Inception network using the
counterfactuals obtained by scaling the input image;
$\pathfn(\sparam) = \sparam~\img$ where $\img$ is the input image.
Similar to the interior gradients, the integrated gradients can also
be aggregated along the color channel to obtain pixel importance
scores which can then be visualized as discussed earlier.
Figure~\ref{fig:intgrad-finalgrad} shows these visualizations
for a bunch of images. For comparison, it also presents the
corresponding visualization obtained from the gradients at the actual
image. From the visualizations, it seems quite evident that the integrated
gradients are better at capturing important features.

\stitle{Attributions are independent of network implementation}
%\stitle{Invariance of attribution}
Two networks may be functionally equivalent\footnote{Formally, two networks
  $F$ and $G$ are functionally equivalent if and only if $\forall~x: F(x) = G(x)$.
}
despite having
very different internal structures. See Figure~\ref{fig:deeplift}
in the appendix for an example.  
Ideally, feature attribution should only be determined by the
functionality of the network and not its implementation.
Attributions generated by integrated gradients satisfy this property
by definition since they are based only on the gradients of the function
represented by the network.

In contrast, this simple property does not hold for all feature attribution
methods known to us, including, DeepLift (\cite{SGSK16}),
Layer-wise relevance propagation (LRP) (\cite{BMBMS16}), Deconvolution networks
(DeConvNets) (\cite{ZF14}), and Guided back-propagation (\cite{SDBR14})
(a counter-example is given in Figure~\ref{fig:deeplift}). We discuss these
methods in more detail in Section~\ref{sec:related}

\subsection{Evaluating Our Approach}
We discuss an evaluation of integrated gradients as a measure of
feature importance, by comparing them against (final) gradients.

\stitle{Pixel ablations}
The first evaluation is based on a method by~\cite{SBMBM15}. Here we
ablate\footnote{Ablation in our setting amounts to zeroing out (or
  blacking out) the intensity for the R, G, B channels. We view this
  as a natural mechanism for removing the information carried by the
  pixel (than, say, randomizing the pixel's intensity as proposed by
  ~\cite{SBMBM15}, especially since the black image is a natural baseline
  for vision
  tasks.}  the top $5000$ pixels ($10\%$ of the image) by importance
score, and compute the score drop for the highest scoring object
class.  The ablation is performed $100$ pixels at a time, in a
sequence of $50$ steps.  At each perturbation step $k$ we measure the
average drop in score up to step $k$. This quantity is referred to a
\emph{area over the perturbation curve} (AOPC) by~\cite{SBMBM15}.

\begin{figure}[!htb]
  \centering
  \includegraphics[width=0.5\textwidth]{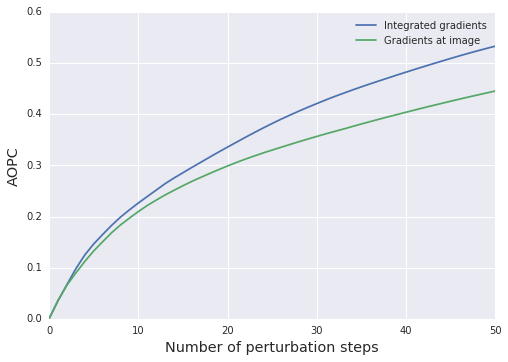}
  \caption{AOPC (\cite{SBMBM15}) for integrated gradients and gradients at image.}\label{fig:aopc}
\end{figure}

Figure~\ref{fig:aopc} shows the AOPC curve with respect to the number
of perturbation steps for integrated gradients and gradients at the
image.  AOPC values at each step represent the average over a dataset
of $150$ randomly chosen images. It is clear that ablating the top
pixels identified by integrated gradients leads to a larger score drop
that those identified by gradients at the image.

Having said that, we note an important issue with the technique.  The
images resulting from pixel perturbation are often unnatural, and it
could be that the scores drop simply because the network has never
seen anything like it in training.

\stitle{Localization}
The second evaluation is to consider images with human-drawn bounding
boxes around objects, and compute the percentage of pixel attribution
inside the bounding box.  We use the 2012 ImageNet object localization
challenge dataset to get a set of human-drawn bounding boxes. We run
our evaluation on $100$ randomly chosen images satisfying the
following properties --- (1) the total size of the bounding box(es) is
less than two thirds of the image size, and (2) ablating the bounding
box significantly drops the prediction score for the object class. (1)
is for ensuring that the boxes are not so large that the bulk of the
attribution falls inside them by definition, and (2) is for ensuring
that the boxed part of the image is indeed responsible for the
prediction score for the image. We find that on $82$ images the
integrated gradients technique leads to a higher fraction of the pixel
attribution inside the box than gradients at the actual image. The
average difference in the percentage pixel attribution inside the box
for the two techniques is $8.4\%$.

%In light of these findings, we propose using integrated gradients for
%object segmentation and localization as a promising future
%direction.
While these results are promising, we note the following
caveat. Integrated gradients are meant to capture pixel importance
with respect to the prediction task. While for most objects, one would
expect the pixel located on the object to be most important for the
prediction, in some cases the context in which the object occurs may
also contribute to the prediction. The “cabbage butterfly” image from
Figure~\ref{fig:intgrad-finalgrad} is a good example of this where the
pixels on the leaf are also surfaced by the integrated gradients.

\stitle{Eyeballing}
Ultimately, it was hard to come up with a perfect evaluation
technique.  So we did spend a large amount of time applying and
eyeballing the results of our technique to various networks---the ones
presented in this paper, as well as some networks used within
products.  For the Inception network, we welcome you to eyeball more
visualizations in Figure~\ref{fig:intgrad-finalgrad-more} in
the appendix and also at: \url{https://github.com/ankurtaly/Attributions}.
While we found our method to beat
gradients at the image for the most part, this is clearly a subjective
process prone to interpretation and cherry-picking, but is also
ultimately the measure of the utility of the approach---debugging
inherently involves the human.

Finally, also note that we did not compare against other whitebox
attribution techniques (e.g., DeepLift (\cite{SGSK16})), because our focus was on black-box
techniques that are easy to implement, so comparing against gradients
seems like a fair comparison.

\subsection{Debugging networks}
Despite the widespread application of deep neural networks to problems
in science and technology, their internal workings largely remain a
black box. As a result, humans have a limited ability to understand
the predictions made by these networks. This is viewed as hindrance in
scenarios where the bar for precision is high, e.g., medical
diagnosis, obstacle detection for robots, etc. (\cite{darpaAI}).
Quantifying feature
importance for individual predictions is a first step towards
understanding the behavior of the network; at the very least, it helps
debug misclassified inputs, and sanity check the internal workings. We
present evidence to support this below.

We use feature importance to debug misclassifications made by the
Inception network. In particular, we consider images from the ImageNet
dataset where the groundtruth label for the image not in the top five
labels predicted by the Inception network. We use interior gradients
to compute pixel importance scores for both the Inception label and
the groundtruth label, and visualize them to gain insight into the cause
for misclassification.

\begin{figure}
  \centering
  \includegraphics[width=0.7\textwidth]{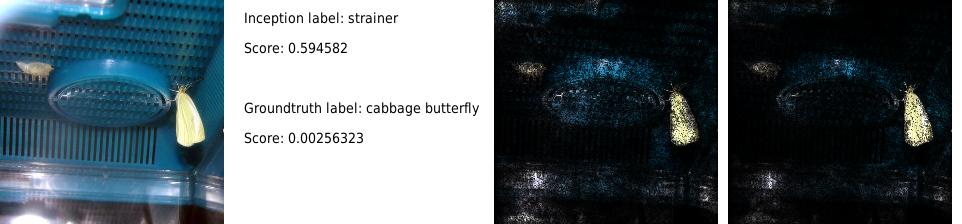}
  \includegraphics[width=0.7\textwidth]{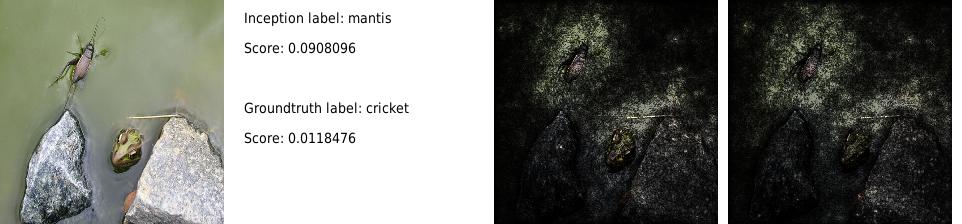}
  \caption{\textbf{Interior gradients of misclassified images.} Left-to-right: Original image,
    Softmax score for the top label assigned by the Inception network and the groundtruth
    label provided by ImageNet, visualization of integrated gradients w.r.t. Inception label,
    visualization of integrated gradients w.r.t. groundtruth label.}\label{fig:attribution-mis}
\end{figure}

Figure~\ref{fig:attribution-mis} shows the visualizations for two misclassified images.
The top image genuinely has two objects, one corresponding to the
groundtruth label and other corresponding to the Inception label. We
find that the interior gradients for each label are able to emphasize
the corresponding objects. Therefore, we suspect that the
misclassification is in the ranking logic for the labels rather than
the recognition logic for each label. For the bottom image, we observe
that the interior gradients are largely similar. Moreover, the cricket
gets emphasized by the interior gradients for the “mantis” (Inception
label). Thus, we suspect this to be a more serious misclassification,
stemming from the recognition logic for the mantis.

\subsection{Discussion}\label{sec:discussion}
\begin{figure}
  \centering
  \includegraphics[width=0.7\textwidth]{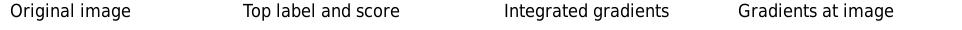}
  \includegraphics[width=0.7\textwidth]{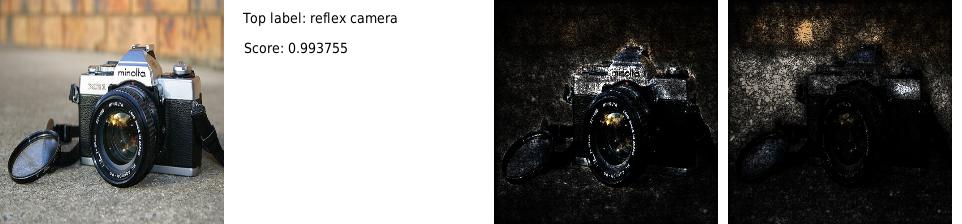}
  \includegraphics[width=0.7\textwidth]{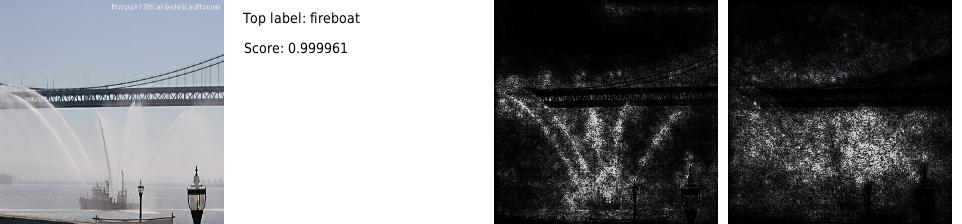}
  \includegraphics[width=0.7\textwidth]{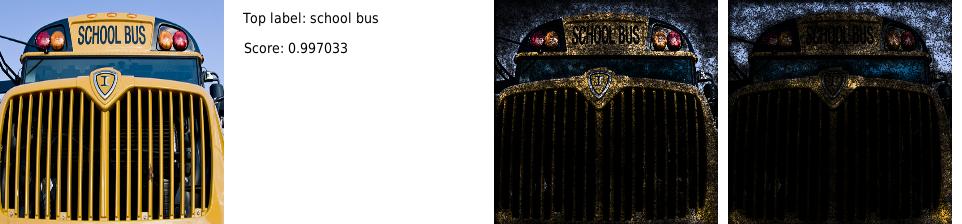}
  \includegraphics[width=0.7\textwidth]{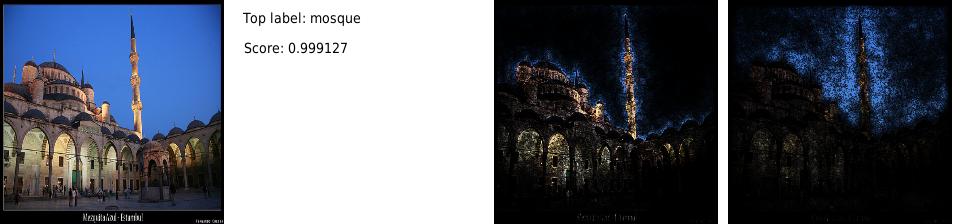}
  \includegraphics[width=0.7\textwidth]{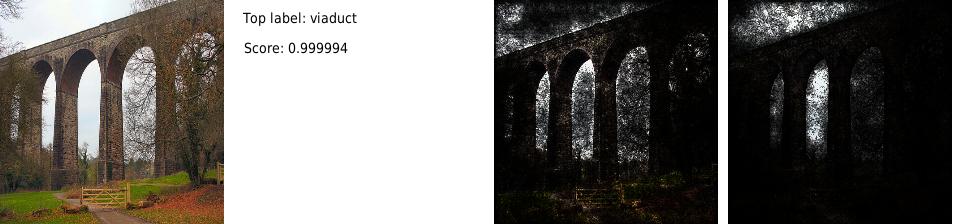}
  \includegraphics[width=0.7\textwidth]{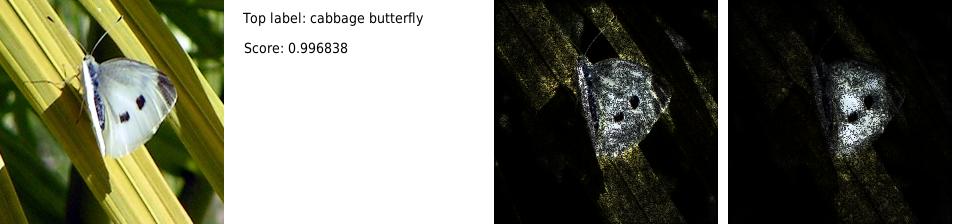}
  \includegraphics[width=0.7\textwidth]{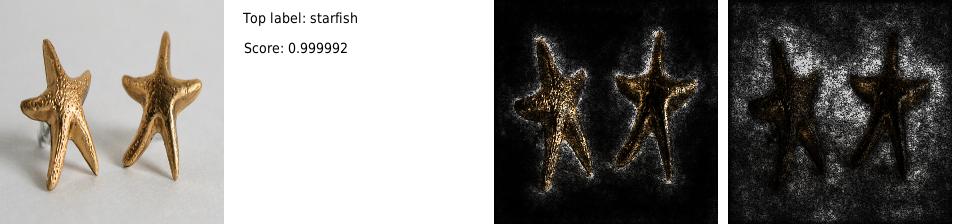}
  \caption{\textbf{Comparing integrated gradients with gradients at the image.}
    Left-to-right: original input image, label and softmax score for
    the highest scoring class, visualization of integrated gradients,
    visualization of gradients at the image. Notice that the visualizations
    obtained from integrated gradients are better at reflecting distinctive
    features of the image.}\label{fig:intgrad-finalgrad}
\end{figure}

\stitle{Faithfullness}
A natural question is to ask why gradients of counterfactuals obtained
by scaling the input capture feature importance for the original image.
First, from
studying the visualizations in Figure~\ref{fig:intgrad-finalgrad},
the results look reasonable in that the highlighted pixels capture features
representative of the predicted class as a human would perceive
them. Second, we confirmed that the network too seems to find these
features representative by performing ablations.
%Specifically, we
%ablate the smallest bounding box drawn around the pixels that are in
%the top $5\%$ by importance score, and observe that the softmax score for
%the class drops significantly.
It is somewhat natural to expect that the Inception network is robust to
to changes in input intensity; presumably there are some low brightness images
in the training set.

However, these counterfactuals seem reasonable even for
networks where such scaling does not correspond to a natural concept
like intensity, and when the counterfactuals fall outside the training
set; for instance in the case of the ligand-based virtual screening network (see
Section~\ref{sec:drug-discovery}).
We \emph{speculate} that the reason why these counterfactuals
make sense is because the network is built by composing ReLUs.
As one scales the input starting from a suitable baseline, various
neurons activate, and the scaling process that does a somewhat thorough
job of exploring all these events that contribute to the prediction for the
input. There is an analogous argument for other operator such as max pool,
average pool, and softmax---here the triggering events aren’t discrete
but the argument is analogous.

%% \stitle{Generation vs. Discrimination}
%% In generative models, all features contribute to single task. In this
%% respect, gradients effectively quantify the importance of each feature
%% to the final prediction score.

%% On the other hand in discriminative models (such ones with a softmax
%% classifier output), features contribute to multiple tasks which are
%% all competing against each other. A feature may be important for task
%% in two ways, either directly by contributing to the prediction for the
%% task, or indirectly by hurting the prediction of other tasks. Features
%% importance quantified by gradients does not distinguish between such
%% direct and indirect contributions, which makes it challenging to
%% interpret.

\stitle{Limitations of Approach}
We discuss some limitations of our technique; in a sense these are
limitations of the problem statement and apply equally to other
techniques that attribute to base input features.
\begin{itemize}
\item \textbf{Inability to capture Feature interactions:} The models could
  perform logic that effectively combines features via a conjunction or
  an implication-like operations; for instance, it could be that a
  molecule binds to a site if it has a certain structure that is
  essentially a conjunction of certain atoms and certain bonds
  between them. Attributions or importance scores have no way to
  represent these interactions.
\item \textbf {Feature correlations:} Feature correlations are a bane to the
  understandability of all machine learning models. If there are two
  features that frequently co-occur, the model is free to assign
  weight to either or both features. The attributions would then respect
  this weight assignment. But, it could be that the specific weight
  assignment chosen by the model is not human-intelligible. Though
  there have been approaches to feature selection that reduce feature
  correlations (\cite{YL03}), it is unclear how they apply to deep models
  on dense input.
\end{itemize}

%% \stitle{When Interior Gradients and Gradients at the image coincide}
%% If the network only consists of RELUs with zero biases, Max/Avg pool nodes,
%% then the function is linear in the scaling of intensities, i.e., all
%% the interior gradients are identical. (Notice that this does not imply
%% that the network is a linear function of its input, just that it is
%% linear to such uniform scaling of intensities.) So it is tempting to
%% quantify feature importance using just the gradient of the input for
%% such networks. But this comes with a caveat. It could still be that
%% the network contrives biases by hiding them in input features. This is
%% easier to see if the network relies on an embedding representation of
%% the input. It could simply use an embedding dimension as a constant
%% and propagate this constant to higher level layers. This would distort
%% the gradient in a way that makes it not reflective of feature
%% importance. Notice this phenomenon could also occur in networks with
%% RELUs that have non zero biases. Although, one would hope that it
%% isn’t as frequent in such networks as the bias variable are available.

\subsection{Related work}\label{sec:related}
Over the last few years, there has been a vast amount work on
demystifying the inner workings of deep networks. Most of this work
has been on networks trained on computer vision tasks, and deals with
understanding what a specific neuron computes (\cite{EBCV09, QVL13}) and
interpreting the representations captured by neurons during a
prediction (\cite{MV15, DB15, YCNFL15}).

Our work instead focuses on understanding the network's behavior on a
specific input in terms of the base level input features. Our
technique quantifies the importance of each feature in the
prediction. Known approaches for accomplishing this can be divided
into three categories.

\stitle{Gradient based methods} The first approach is to use gradients
of the input features to quantify feature importance (\cite{BSHKHM10,
  SVZ13}). This approach is the easiest to implement. However, as
discussed earlier, naively using the gradients at the actual input does
not accurate quantify feature importance as gradients suffer from
saturation.

\stitle{Score back-propagation based methods} The second set of
approaches involve back-propagating the final prediction score through
each layer of the network down to the individual features.  These
include DeepLift (\cite{SGSK16}), Layer-wise relevance propagation (LRP)
 (\cite{BMBMS16}), Deconvolution networks (DeConvNets) (\cite{ZF14}), and
Guided back-propagation (\cite{SDBR14}). These methods largely differ in
the backpropagation logic for various non-linear activation
functions. While DeConvNets, Guided back-propagation and LRP rely on
the local gradients at each non-linear activation function, DeepLift
relies on the deviation in the neuron's activation from a certain
baseline input.

Similar to integrated gradients, the DeepLift and LRP also result in
an exact distribution of the prediction score to the input
features. However, as shown by Figure~\ref{fig:deeplift}, the
attributions are not invariant across functionally equivalent
networks. Besides, the primary advantage of our method over all these
methods is its ease of implementation.  The aforesaid methods require
knowledge of the network architecture and the internal neuron
activations for the input, and involve implementing a somewhat
complicated back-propagation logic. On the other hand, our method is
agnostic to the network architectures and relies only on computing
gradients which can done efficiently in most deep learning frameworks.

\stitle{Model approximation based methods} The third approach, proposed
first by~\cite{RSG16a, RSG16b}, is to
locally approximate the behavior of the network in the vicinity of the
input being explained with a simpler, more interpretable model.
An appealing aspect of this approach is that it is
completely agnostic to the structure of the network and only deals
with its input-output behavior. The approximation is learned by
sampling the network's output in the vicinity of the input at hand. In
this sense, it is similar to our approach of using
counterfactuals. Since the counterfactuals are chosen somewhat
arbitrarily, and the approximation is based purely on the network's
output at the counterfactuals, the faithfullness question is far more
crucial in this setting. The method is also expensive to implement as
it requires training a new model locally around the input being explained.

\section{Applications to Other Networks}\label{sec:beyond-inception}
The technique of quantifying feature importance by inspecting
gradients of counterfactual inputs is generally applicable across deep
networks. While for networks performing vision tasks, the counterfactual
inputs are obtained by scaling pixel intensities, for other
networks they may be obtained by scaling an embedding representation
of the input.

As a proof of concept, we apply the technique to the molecular graph
convolutions network of~\cite{KMBPR16} for ligand-based virtual
screening and an LSTM model (\cite{ZSV14}) for the language modeling
of the Penn Treebank dataset (\cite{MSM94}).

\subsection{Ligand-Based Virtual Screening}\label{sec:drug-discovery}
The Ligand-Based Virtual Screening problem is to predict whether an
input molecule is active against a certain target (e.g., protein or
enzyme). The process is meant to aid the discovery of new drug
molecules. Deep networks built using molecular graph convolutions have
recently been proposed by~\cite{KMBPR16} for solving this problem.

Once a molecule has been identified as active against a target, the
next step for medicinal chemists is to identify the molecular
features---formally, \emph{pharmacophores}\footnote{A pharmacophore is
  the ensemble of steric and electronic features that is necessary to
  ensure the a molecule is active against a specific biological target
  to trigger (or to block) its biological response.}---that are
responsible for the activity.  This is akin to quantifying feature
importance, and can be achieved using the method of integrated
gradients. The attributions obtained from the method help with
identifying the dominant molecular features, and also help sanity
check the behavior of the network by shedding light on its inner
workings. With regard to the latter, we discuss an anecdote later in
this section on how attributions surfaced an anomaly in W1N2 network
architecture proposed by~\cite{KMBPR16}.

%The paper \cite{KMBPR16} specifies a
%taxonomy of network architectures for convolving the atom and atom-pair features.
%varying convolution depths. While our technique applies to all architectures,
%below we discuss our results from applying it to the W2N2 architecture which
%contains two convolutional layers and only considers atom pairs upto a maximum
%graph distance of two.

\stitle{Defining the counterfactual inputs} The first step in computing
integrated gradients is to define the set of counterfactual inputs. The
network requires an input molecule to be encoded by hand as a set of
atom and atom-pair features describing the molecule as an undirected
graph. Atoms are featurized using a one-hot encoding specifying the atom
type (e.g., C, O, S, etc.), and atom-pairs are featurized by specifying either the type
of bond (e.g., single, double, triple, etc.) between the atoms, or the
graph distance between them
\footnote{This featurization is referred to as ``simple'' input featurization
  in~\cite{KMBPR16}.}

The counterfactual inputs are obtained by scaling down the molecule
features down to zero vectors, i.e., the set $\{\sparam
\molfeatures(\mol)~\vert~0 \leq \sparam \leq 1\}$ where
$\molfeatures(mol)$ is an encoding of the molecule into atom and
atom-pair features.

The careful reader might notice that these counterfactual inputs are
not valid featurizations of molecules. However, we argue that they are
still valid inputs for the network. First, all opera- tors in the
network (e.g., ReLUs, Linear filters, etc.) treat their inputs as
continuous real numbers rather than discrete zeros and ones. Second,
all fields of the counterfactual inputs are bounded be- tween zero and
one, therefore, we don't expect them to appear spurious to the
network. We discuss this further in section~\ref{sec:discussion}

In what follows, we discuss the behavior of a network based on the
W2N2-simple architecture proposed by~\cite{KMBPR16}.  On inspecting
the behavior of
the network over counterfactual inputs, we observe saturation here as
well. Figure~\ref{fig:sat-gas-softmax} shows the trend in the softmax
score for the task PCBA-588342 for twenty five active molecules as we
vary the scaling parameter $\sparam$ from zero to one. While the
overall saturated region is small, saturation does exist near vicinity
of the input ($0.9 \leq \sparam \leq
1$). Figure~\ref{fig:sat-gas-features} in the appendix shows that the
total feature gradient varies significantly along the scaling path;
thus, just the gradients at the molecule is fully indicative of the
behavior of the network.

\begin{figure}
  \centering
  \includegraphics[width=0.9\textwidth]{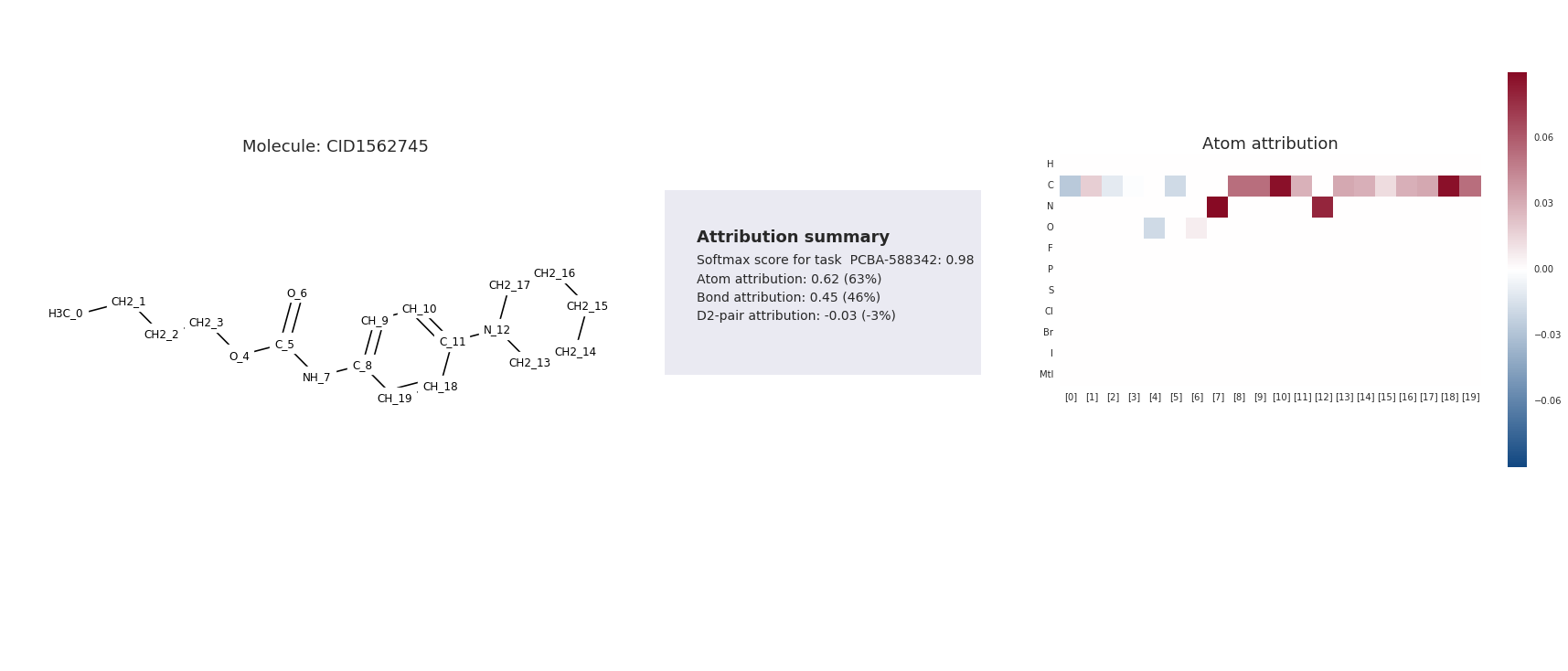}
  \includegraphics[width=0.9\textwidth]{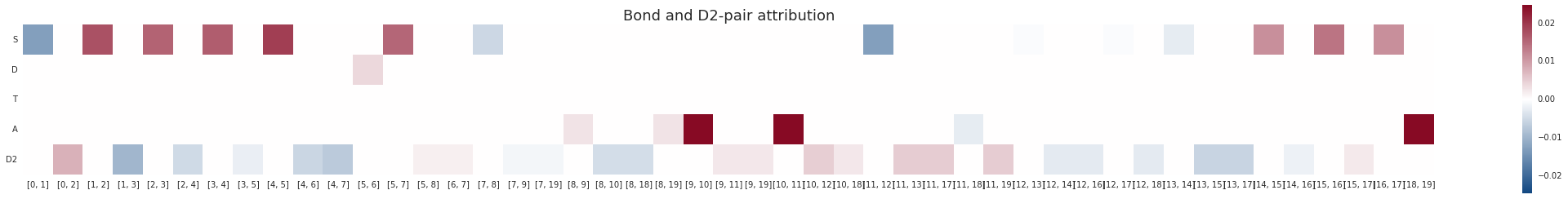}
  \caption{\textbf{Attribution for a molecule under the W2N2 network (\cite{KMBPR16})}.
    The molecules is active on task PCBA-58432.}\label{fig:attribution-gas}
\end{figure}

\stitle{Visualizing integrated gradients} We cumulate the gradients
of these counterfactual inputs to obtain an attribution of the
prediction score to each atom and atom-pair feature. Unlike image
inputs, which have dense features, the set of input features for
molecules are sparse.  Consequently, the attributions are sparse and
can be inspected directly.  Figure~\ref{fig:attribution-gas}
shows heatmaps for the atom and atom-pair attributions for a specific
molecule.

Using the attributions, one can easily
identify the atoms and atom-pairs that that have a strongly positive
or strongly negative contribution. Since the attributions add up to
the final prediction score (see Proposition~\ref{prop:additivity}),
the attribution magnitudes can be use for accounting the
contributions of each feature. For instance, the atom-pairs that have
a bond between them contribute cumulatively contribute $46\%$ of the
prediction score, while all other atom pairs cumulatively contribute
$-3\%$.

We presented the attributions for $100$ molecules active against
a specific task to a few chemists. The chemists were able to immediately
spot dominant functional groups (e.g., aromatic rings) being surfaced
by the attributions. A next step could be cluster the aggregate the
attributions across a large set of molecules active against a specific
task to identify a common denominator of features shared by all
active molecules.

\stitle{Identifying Dead Features}
We now discuss how attributions helped us spot an anomaly in the W1N2
architecture. On applying the integrated gradients method to the W1N2
network, we found that several atoms in the same molecule received the
exact same attribution. For instance, for the molecule in
Figure~\ref{fig:attribution-gas}, we found that several carbon atoms
at positions $2$, $3$, $14$, $15$, and $16$ received the same
attribution of $0.0043$ despite being bonded to different atoms, for
e.g., Carbon at position $3$ is bonded to an Oxygen whereas Carbon at
position $2$ is not. This is surprising as one would expect two atoms
with different neighborhoods to be treated differently by
the network.

On investigating the problem further we found that since the W1N2
network had only one convolution layer, the atoms and atom-pair
features were not fully convolved. This caused all atoms that have the
same atom type, and same number of bonds of each type to contribute
identically to the network. This is not the case for
networks that have two or more convolutional layers.

Despite the aforementioned problem, the W1N2 network had good
predictive accuracy. One hypothesis for this is that the atom type
and their neighborhoods are tightly correlated; for instance an
outgoing double bond from a Carbon is always to another Carbon or
Oxygen atom. As a result, given the atom type, an explicit encoding
of the neighborhood is not needed by the network. This also
suggests that equivalent predictive accuracy can be achieved using a
simpler ``bag of atoms'' type model.

\subsection{Language Modeling}\label{sec:lm}

To apply our technique for language modeling,
we study word-level language modeling of the Penn Treebank dataset (\cite{MSM94}),
and apply an LSTM-based sequence model based on~\cite{ZSV14}.
For such a network, given a sequence of input words, and the softmax prediction for
the next word,
we want to identify the importance of the preceding words for the score.

\begin{figure}[!htb]
  \centering
  \includegraphics[width=0.9\textwidth]{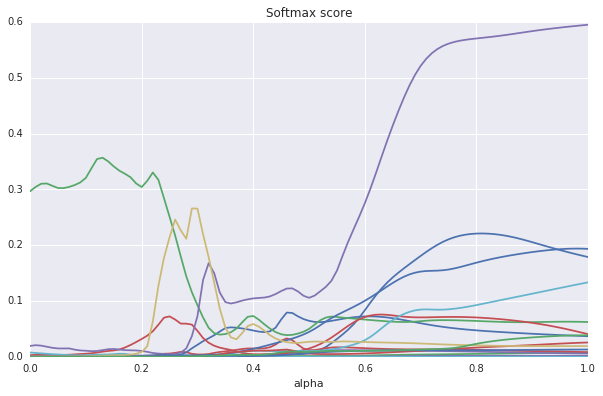}
  \caption{Softmax scores of the next word in the LSTM language model (Section~\ref{sec:lm})}\label{fig:ptb}
\end{figure}

As in the case of the Inception model, we observe saturation in this LSTM network.
To describe the setup, we choose 20 randomly chosen sections of the test data,
and for each of them inspect the prediction score of the next word using the
first 10 words.
Then we give each of the 10 input words a weight of $\alpha\in[0,1]$,
which is applied to scale their embedding vectors.
In Figure~\ref{fig:ptb}, we plot the prediction score as a function of $\alpha$.
For all except one curves, the curve starts near zero at $\alpha=0$, moves around
in the middle, stabilizes, and turns flat around $\alpha=1$.
For the interesting special case where softmax score is non-zero at $\alpha=0$,
it turns out that that the word being predicted represents
out of vocabulary words.
[!h]
\begin{figure}
  \small
\begin{center}
  \begin{tabular}{ | l | c c c c c  r |}
    \hline
    Sentence & the & shareholders & claimed & more & \textbf{than} & \$ N millions in losses \\ \hline
    Integrated gradients & -0.1814& -0.1363 & 0.1890& {\bf 0.6609} &  &  \\ 
    Gradients & 0.0007 & -0.0021 & 0.0054 & -0.0009 &  &  \\ \hline
  \end{tabular}
\end{center}
\caption{Prediction for \textbf{than}: 0.5307, total integrated gradient: 0.5322}
\label{tab:ptb_near}
\end{figure}

\begin{figure}[!h]
  \small
\begin{center}
  \begin{tabular}{ | l | c c c c c c r |}
    \hline
    Sentence               & and       & N   & minutes  & after  &   the    & ual    & trading  \\ \hline
    Integrated gradients (*1e-3) & 0.0707&0.1286& 0.3619& 1.9796& -0.0063& {\bf 4.1565}& 0.2213\\ 
    Gradients (*1e-3)          & 0.0066& 0.0009& 0.0075& 0.0678& 0.0033& 0.0474& 0.0184 \\ \hline
    Sentence (Cont.)       & halt      & came      & news      &that       &the        &\textbf{ual}& group  \\ \hline
    Integrated gradients (*1e-3) & -0.8501& -0.4271& 0.4401& -0.0919& 0.3042 &  &  \\ 
    Gradients (*1e-3) & -0.0590 &  -0.0059& 0.0511& 0.0041&0.0349 &  &  \\ \hline
  \end{tabular}
\end{center}
\caption{Prediction for \textbf{ual}: 0.0062, total integrated gradient: 0.0063}
\label{tab:ptb_far}
\end{figure}

In Table~\ref{tab:ptb_near} and Table~\ref{tab:ptb_far} we show two comparisons of gradients to integrated gradients.
Due to saturation, the magnitudes of gradients are so small compared to the
prediction scores that it is difficult to make sense of them.
In comparison, (approximate) integrated gradients have a total amount close to the prediction,
and seem to make sense. For example, in the first example, the integrated gradients attribute the prediction score of ``than``
to the preceding word ``more''. This makes sense as ``than'' often follows right after ``more`` in English.
On the other hand, standard gradient gives a slightly negative attribution that betrays our intuition.
In the second example, in predicting the second ``ual'', integrated gradients are clearly
the highest for the first occurrence of ``ual'', which is the only word that is highly predictive of the second ``ual''.
On the other hand, standard gradients are not only tiny, but also similar in magnitude for multiple words.

%% In
%% contrast, our method is extremely easy to implement, and is faithful
%% to the original network at least in offering an exact attribution of
%% the network's output to the input features.

\section{Conclusion}
We present Interior Gradients, a method for quantifying feature
importance. The method can be applied to a variety of deep networks
without instrumenting the network, in fact, the amount of code
required is fairly tiny.  We demonstrate that it is possible to have
some understanding of the performance of the network without a
detailed understanding of its implementation, opening up the
possibility of easy and wide application, and lowering the bar on the
effort needed to debug deep networks.

We also wonder if Interior Gradients are useful within training as a
measure against saturation, or indeed in other places that gradients
are used.

\subsubsection*{Acknowledgments}
We would like to thank Patrick Riley and Christian Szegedy for their
helpful feedback on the technique and on drafts of this paper. 

\bibliography{paper-iclr.bib}
\bibliographystyle{iclr2017_conference}
\newpage
\appendix
\section{Appendix}

\begin{figure}[!htb]
  \centering
  \includegraphics[width=0.7\textwidth]{Figures/IntegratedGrads/img0.jpg}
  \includegraphics[width=0.7\textwidth]{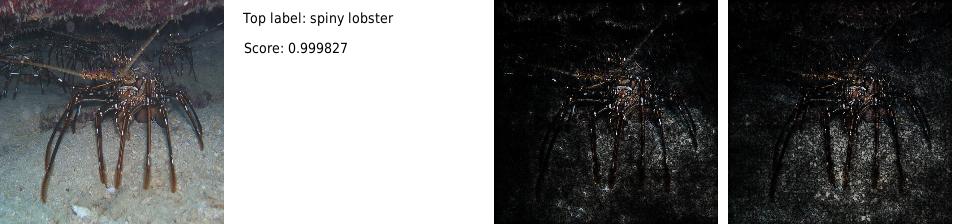}
  \includegraphics[width=0.7\textwidth]{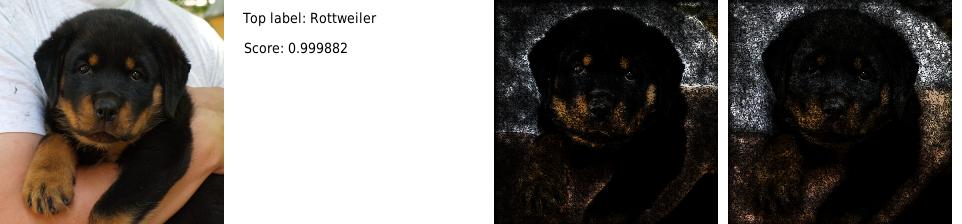}
  \includegraphics[width=0.7\textwidth]{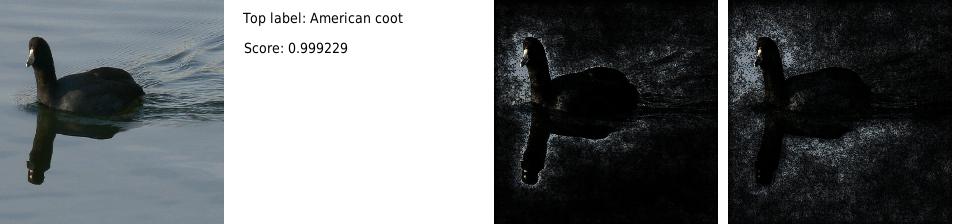}
  \includegraphics[width=0.7\textwidth]{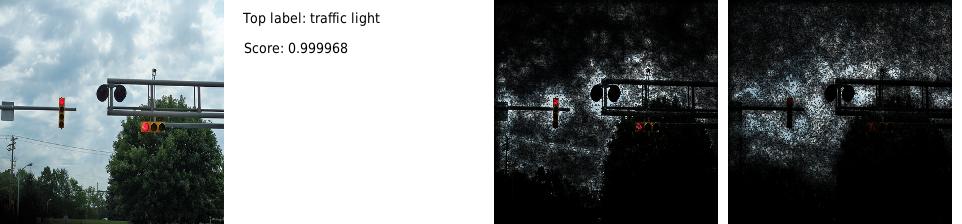}
  \includegraphics[width=0.7\textwidth]{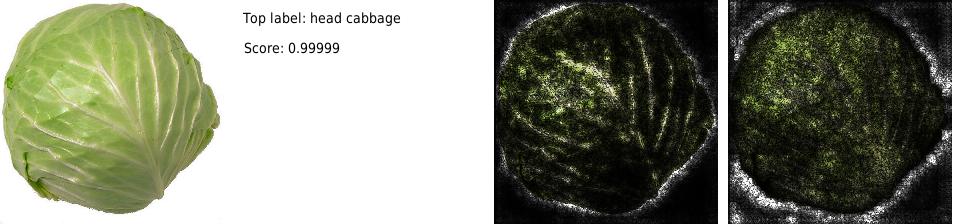}
  \includegraphics[width=0.7\textwidth]{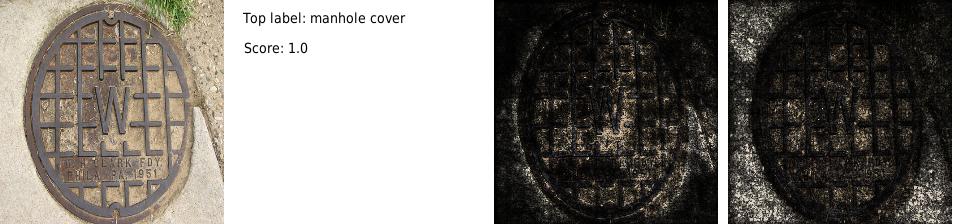}
  \includegraphics[width=0.7\textwidth]{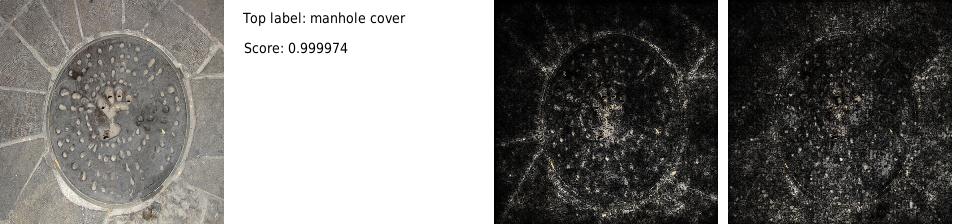}
  \includegraphics[width=0.7\textwidth]{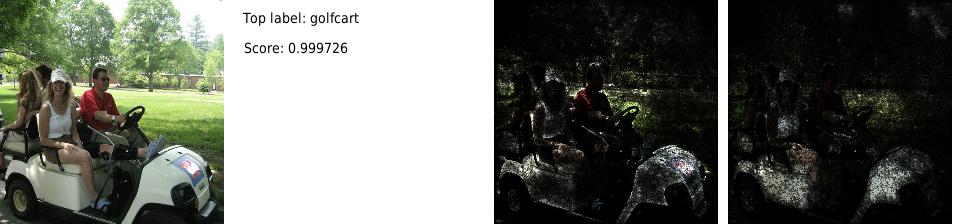}
  \caption{\textbf{More visualizations comparing integrated gradients with gradients at the image.}
    Left-to-right: original input image, label and softmax score for
    the highest scoring class, visualization of integrated gradients,
    visualization of gradients at the image.}\label{fig:intgrad-finalgrad-more}
\end{figure}

\begin{figure}
  \centering
  \begin{subfigure}{.4\textwidth}
    \centering
    \includegraphics[width=0.9\textwidth]{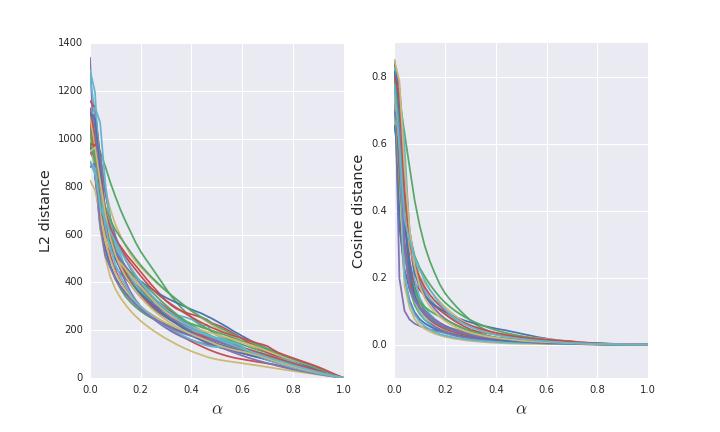}
    \caption*{Layer {\tt mixed5b}}
  \end{subfigure}%
  \begin{subfigure}{.4\textwidth}
    \centering
    \includegraphics[width=0.9\textwidth]{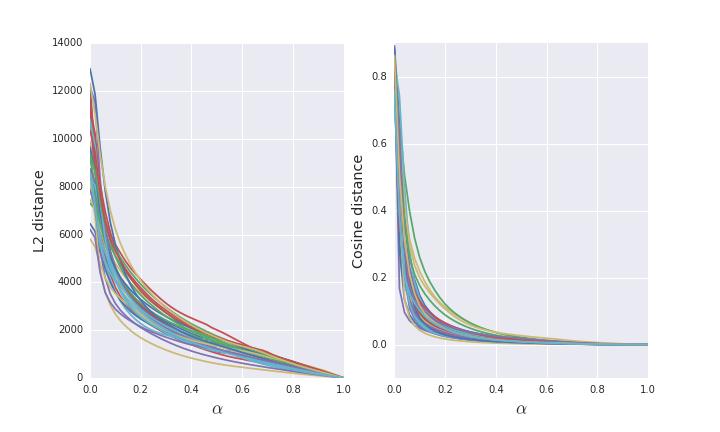}
    \caption*{Layer {\tt mixed4d}}
  \end{subfigure}
  \begin{subfigure}{.4\textwidth}
    \centering
    \includegraphics[width=0.9\textwidth]{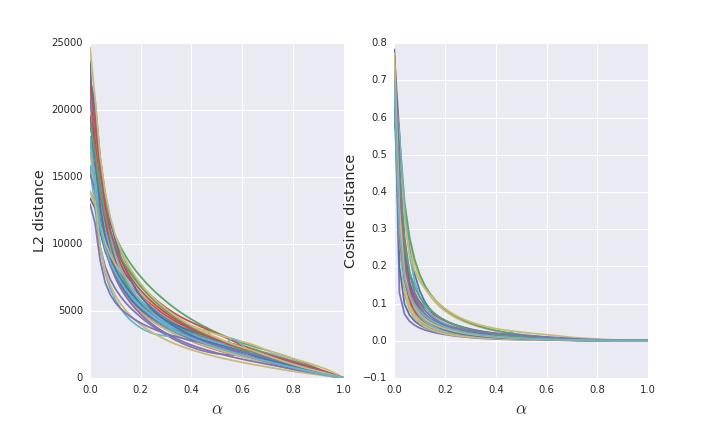}
    \caption*{Layer {\tt mixed4b}}
  \end{subfigure}%
  \begin{subfigure}{.4\textwidth}
    \centering
    \includegraphics[width=0.9\textwidth]{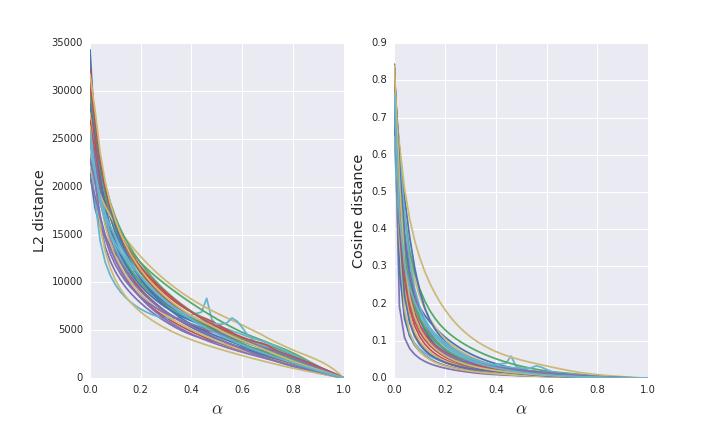}
    \caption*{Layer {\tt mixed3b}}
  \end{subfigure}
  \caption{\textbf{Saturation in intermediate layers of Inception.} For each layer we plot the
    L2 and Cosine distance between the activation vector for a scaled down
    image and the actual input image, with respect to the scaling parameter.
    Each plot shows the trend for 30 randomly chosen images from the ImageNet
    dataset. Notice that trends in all plots flatten as the scaling parameter
    increases. For the deepest Inception layer {\tt mixed5b}, the Cosine distance
    to the activation vector at the image is less than $0.01$ when $\sparam > 0.6$,
    which is really tiny given that this layer has $50176$ neurons.}\label{fig:sat-incp-intermediate}
\end{figure}

\begin{figure}
    \begin{center}
    \begin{subfigure}{.5\textwidth}
      \centering
      \includegraphics[width=0.9\textwidth]{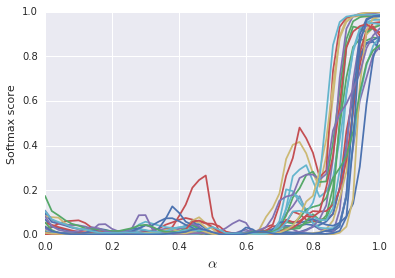}
      \caption{Softmax score for task}\label{fig:sat-gas-softmax}
    \end{subfigure}%
    \begin{subfigure}{.5\textwidth}
      \centering
      \includegraphics[width=0.9\textwidth]{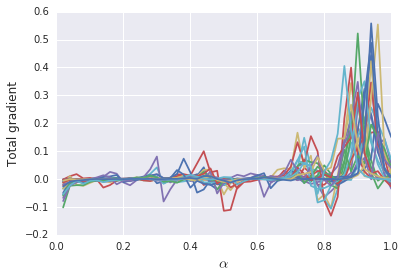}
      \caption{Sum of the feature gradients}\label{fig:sat-gas-features}
    \end{subfigure}
  %\fbox{\rule[-.5cm]{0cm}{4cm} \rule[-.5cm]{4cm}{0cm}}
  \end{center}
    \caption{\textbf{Saturation in the W2N2 network (\cite{KMBPR16})}. Plots for the
      softmax score for task PCBA-58834, and the sum of the feature gradients w.r.t. the same task
      for twenty molecules. All molecules are active against the task}\label{fig:sat-gas}
\end{figure}

\begin{figure}
  \centering
  \begin{subfigure}{.75\textwidth}
    \includegraphics[width=\textwidth]{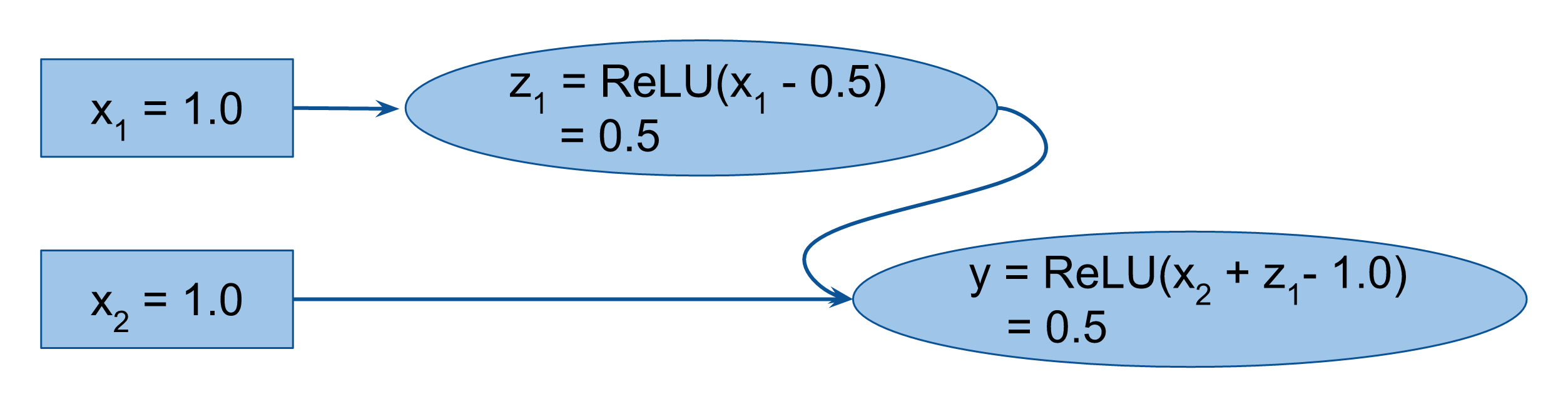}    
    \caption{
      $\begin{array}{ll}
        \mbox{Integrated gradients} & x_1 = 0.25,~x_2 = 0.25 \\
        \mbox{DeepLift} & x_1 = \tfrac{0.5}{3},~x_2 = \tfrac{1}{3} \\
        \mbox{Layer-wise relevance propagation} & x_1 = \tfrac{0.5}{3},~x_2 = \tfrac{1}{3} \\
        \mbox{DeConvNet} & x_1 = 1.25,~x_2 = 0.75 \\
        \mbox{Guided backpropagation} & x_1 = 1.25,~x_2 = 0.75 \\
       \end{array}$
    }\label{fig:deeplift-1}
  \end{subfigure}
  \begin{subfigure}{.85\textwidth}
    \includegraphics[width=\textwidth]{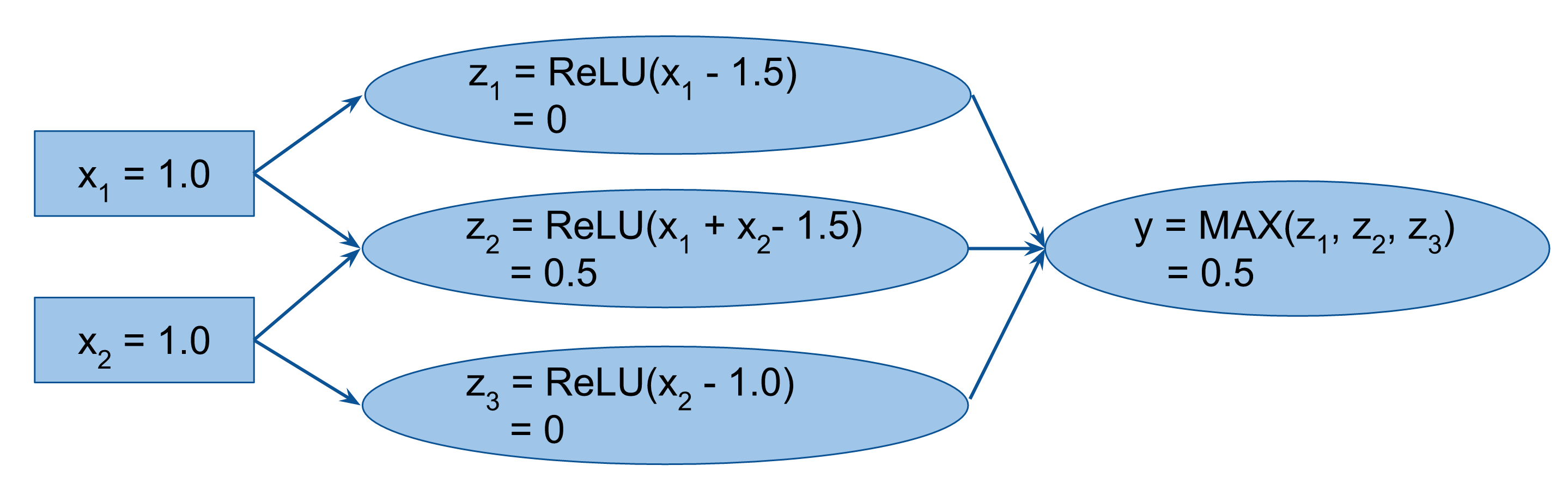}
    \caption{
      $\begin{array}{ll}
        \mbox{Integrated gradients} & x_1 = 0.25,~x_2 = 0.25 \\
        \mbox{DeepLift} &  x_1 = 0.25,~x_2 = 0.25 \\
        \mbox{Layer-wise relevance propagation} & x_1 = 0.25,~x_2 = 0.25 \\
        \mbox{DeConvNet} &  x_1 = 1.0,~x_2 = 1.0 \\
        \mbox{Guided backpropagation} & x_1 = 1.0,~x_2 = 1.0 \\
       \end{array}$
    }\label{fig:deeplift-2}
  \end{subfigure}
  \caption{\textbf{Attributions for two functionally equivalent networks} using
    integrated gradients, DeepLift (\cite{SGSK16}), Layer-wise relevance propagation
    (\cite{BMBMS16}), DeConvNets (\cite{ZF14}), and Guided backpropagation (\cite{SDBR14}).
    The input is $x_1 = 1.0,~x_2 = 1.0$. The reference input
    for the DeepLift method is $x_1 = 0,~x_2 = 0$. All methods except integrated
    gradients provide different attributions for the two networks.}\label{fig:deeplift}
\end{figure}

\end{document}